%% file: root.tex
\documentclass[letterpaper, 10 pt, conference]{ieeeconf}  

\usepackage{makecell}
\usepackage{float}
\usepackage{tabularx} 
\usepackage{graphicx}
\usepackage{amsmath}
\usepackage{bbding}
\usepackage{pifont}
\usepackage{wasysym}
\usepackage{amssymb}
\usepackage{hyperref}
\usepackage{threeparttable} 
\usepackage{cite} 
\usepackage{booktabs}
\usepackage{xcolor}
\usepackage{multirow}
\usepackage{blindtext}
\usepackage{flushend} 
\hypersetup{hidelinks}   
\usepackage{bm}
\usepackage{subcaption}
\usepackage[linesnumbered, ruled]{algorithm2e}
\usepackage[normalem]{ulem}

\SetKwInput{KwInput}{Input}
\SetKwInput{KwOutput}{Output}

\IEEEoverridecommandlockouts                              

\title{\LARGE \bf
DriveCode: Domain Specific Numerical Encoding for 
\\LLM-Based Autonomous Driving
}

\author{Zhiye Wang$^{2\ast\ddagger}$, Yanbo Jiang$^{1\ast}$, Rui Zhou$^{2}$, Bo Zhang$^{3,4}$, \\ Fang Zhang$^{5\dagger}$, Zhenhua Xu$^{1\dagger}$, Yaqin Zhang$^{3}$, Jianqiang Wang$^{1,5}$%
 \thanks{This work is supported by the National Natural Science Foundation of China (No. 52221005), and Tsinghua University-Toyota Motor Corporation Joint Research Center for Al Technology Automated Vehicle.}
 \thanks{$^{1}$The School of Vehicle and Mobility, Tsinghua University, Beijing, China}%
 \thanks{$^{2}$School of Information Science and Engineering, Lanzhou University, Lanzhou, China}%
          \thanks{$^{3}$The Institute for AI Industry Research (AIR), Tsinghua University, Beijing, China.
         }%
          \thanks{$^{4}$DiDi, Beijing, China
         }%
         \thanks{$^{5}$State Key Laboratory of Intelligent Green Vehicle and Mobility, Tsinghua University, Beijing, China
         }%
         \thanks{$^{\ast}$\textit{These authors contributed equally to this work.}}
 \thanks{$^{\dagger}$\textit{Corresponding author: Zhenhua Xu, Fang Zhang}}
 \thanks{$^{\ddagger}$
\parbox[t]{0.95\linewidth}{
\textit{Work done during an internship at the School of Vehicle and Mobility, Tsinghua University.}
}}
}

\begin{document}

\maketitle
\enlargethispage{-\baselineskip}
\thispagestyle{empty}
\pagestyle{empty}

\begin{abstract}

Large language models (LLMs) have shown great promise for autonomous driving. However, discretizing numbers into tokens
may weaken precise numerical modeling, 
rely mainly on positional encoding to represent digit place values, and can make it challenging to balance decoding efficiency with numerical precision.
These limitations affect both the processing of sensor measurements and the generation of precise control commands, posing a practical challenge for LLM-based autonomous driving systems that require accurate numerical prediction. 
In this paper, we introduce DriveCode, a novel numerical encoding method that represents numbers as dedicated embeddings rather than discrete text tokens. DriveCode employs a number projector to map numbers into the language model's hidden space, enabling seamless integration with visual and textual features in a unified multimodal sequence. Evaluated on OmniDrive, DriveGPT4, and DriveGPT4-V2 datasets, DriveCode demonstrates superior performance in trajectory prediction and control signal generation, confirming its effectiveness for LLM-based autonomous driving systems. The webpage of this paper is available at \url{https://shiftwilliam.github.io/DriveCode}.
\end{abstract}

\section{Introduction}

Autonomous driving, as an interdisciplinary field of artificial intelligence and transportation, has made remarkable progress in recent years. Typically, autonomous vehicles employ complex modular architectures that encompass perception, planning, and control \cite{liu2021role}. 
These components often depend on extensive manual design and domain knowledge to handle diverse driving scenarios, which leads to high system complexity and substantial integration costs.\par

To address these challenges, end-to-end autonomous driving is emerging as a key research direction in this field \cite{jaeger2023hidden,jiang2023vad,hu2023planning,xu2025drivegpt4}. The core objective of end-to-end autonomous driving is to map sensory inputs, such as image data collected by cameras and distance data obtained from LiDARs and radars, directly to driving commands (e.g., steering angle, throttle position, and braking intensity) through a single model. In recent years, large language models (LLMs) have demonstrated significant potential across multiple domains, including text generation \cite{zhang2024tinyllama,chiang2023vicuna}, image understanding \cite{liu2023llava,bai2025qwen2}, and video analysis \cite{lin2023video,liu2024llavanext}, showcasing their versatility and robust reasoning capabilities. These capabilities make LLMs particularly suited for addressing complex decision-making requirements for end-to-end autonomous driving. Integrating LLMs' reasoning with end-to-end autonomous driving has emerged as a promising direction toward more intelligent autonomous vehicles.\par

However, despite their strong language modeling capabilities, LLMs may still struggle to consistently represent numerical semantics under standard token-based modeling. 
One reason is that numbers are tokenized as text fragments rather than represented as quantitative values. Since decimal points and digit positions are not explicitly modeled in terms of their actual numerical magnitude, the model fails to reliably capture numerical values, leading to errors in numerical comparison and calculation.\par

\begin{figure}[t]
    \centering
    \includegraphics[width=1\linewidth,trim=2cm 2.6cm 2cm 2cm,
  clip]{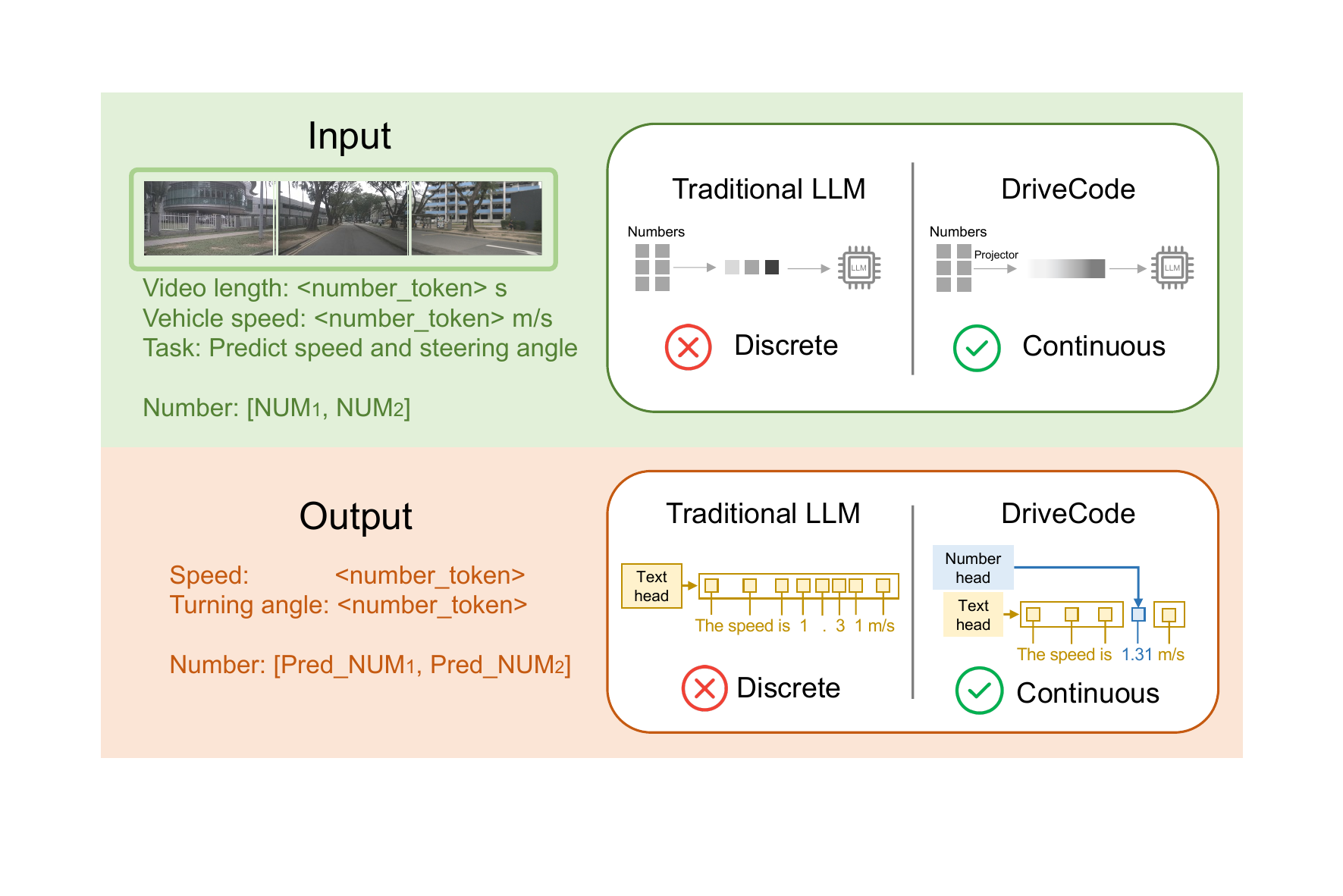}
    \caption{A sample procedure of DriveCode. Numbers are first extracted from text prompts and then processed by a number projector to achieve continuous number processing.}
    \label{fig:brief_intruction}
    
\end{figure} 
 
In the field of autonomous driving, accurate numerical processing is critical, as vehicle control depends on continuous physical quantities such as speed, acceleration, and steering angle. Even small numerical errors can propagate through perception, planning, and control modules, potentially causing unstable trajectories or unsafe maneuvers. 
Notably, autonomous driving systems and LLMs handle numerical errors differently: the former are sensitive to absolute deviations in physical units, while the latter primarily capture token-level differences rather than numerical magnitude. Therefore, standard textual tokenization can introduce a practical mismatch for LLM-based autonomous driving systems.\par

In this paper, we introduce DriveCode, a novel numerical encoding scheme
designed for LLM-based autonomous driving. 
DriveCode processes a sequence of video frames from RGB cameras as input to predict the vehicle's control signals, including speeds, waypoints, and steering angles. 
Instead of converting numbers into discrete text tokens, DriveCode processes them as continuous values throughout: on the input side, a number projector maps each number into the language model's hidden space alongside textual and visual features; on the output side, an LM head and a number head work together, enabling the model to produce both texts and numbers simultaneously rather than generating numbers digit by digit.
This cross-modal alignment enables the model to reason over numbers with higher precision than conventional discrete tokenization, supporting more accurate mapping from perception to control. Fig. \ref{fig:brief_intruction} provides a simplified illustration of DriveCode.

The contributions of this paper are summarized as follows:
\begin{enumerate}


\item We design a number projector that maps numbers into the language model's hidden space as a dedicated modality, enabling them to be jointly processed with textual and visual features rather than treated as discrete text tokens.

\item We introduce a number head that directly regresses numbers from hidden states, allowing the model to produce both natural language responses and precise numerical predictions within a single output sequence.

\item We evaluate DriveCode across multiple autonomous driving datasets, comparing it against baseline methods and conducting ablation studies. The results demonstrate superior performance, confirming the effectiveness of DriveCode in enhancing LLM-based autonomous driving systems.
\end{enumerate}
\section{Related Work}
\subsection{Multimodal Large Language Models}
The rise of Multimodal Large Language Models (MLLMs) has become a transformative force in the field of artificial intelligence, enabling machines to process and generate content across multiple modalities, such as text \cite{zeng2025glm,kassianik2025llama,zhao2025benchmarking}, images \cite{zhang2025llava,cocchi2025llava,liu2023llava}, audio \cite{rouditchenko2025omni,florea2025exploring}, and video\cite{li2024surveying,huang2024vtimellm}. Vision Transformer (ViT) \cite{dosovitskiy2020image} revolutionized the field of computer vision by introducing the Transformer architecture \cite{vaswani2017attention} to visual tasks. BLIP \cite{li2022blip} is a significant contribution to the field of vision-language models, aiming to bridge the gap between visual and language modalities. 
More recently, LLaVA \cite{liu2023llava} demonstrated that large language models can serve as a unified multimodal reasoning core by aligning visual features with LLMs through lightweight projection modules. 
LLaVA-NeXT \cite{liu2024llavanext} further enhanced this paradigm by improving visual understanding and reasoning capacity, establishing a strong foundation for MLLMs' applications.


\subsection{End-to-End Autonomous Driving}
End-to-end driving is a promising paradigm as it circumvents the drawbacks associated with modular systems \cite{hu2023planning,jiang2023vad,chen2024vadv2,shao2024lmdrive,xu2024drivegpt4,xu2025drivegpt4,xu2024insmapper,zhang2025invdriver}, such as their overwhelming complexity and propensity for error propagation \cite{chib2023recent}. DriveGPT4 \cite{xu2024drivegpt4} is one of the pioneering works that leverages large language models for interpretable end-to-end autonomous driving, alleviating the black-box nature of conventional deep models. Subsequent studies have explored interpretable LLM-based trajectory prediction, including LC-LLM \cite{peng2025lc} for lane-change intention prediction and SAM-LLM \cite{cao2025sam} for parametric lane-change reasoning. AutoVLA \cite{zhou2025autovla} further unifies reasoning and action generation by integrating fast trajectory-only planning with slow chain-of-thought reasoning to improve planning efficiency.
Nevertheless, ensuring reliability and mitigating hallucinations in safety-critical driving scenarios remain open challenges \cite{li2025applications,zhu2025survey,wang2024hallucination}. Although recent works such as SimLingo~\cite{renz2025simlingo} explicitly model numerical inputs and action outputs, they rely on fixed and task-specific numerical expression formats for multimodal processing. In contrast, our text-grounded numerical interface directly incorporates numerical inputs alongside language, enabling flexible combinations of numerical and textual information while remaining seamlessly compatible with existing LLMs without requiring specialized architectural modifications or restricted input schemas.

\subsection{Numerical Encoding Methods}
In recent years, substantial efforts have been devoted to improving the numerical understanding capabilities of large language models \cite{lee2023teaching,shen2023positional,schwartz2024numerologic,zhang2025safeauto,golkar2023xval,alberts2024interleaving}. In parallel, theoretical analyses have examined how LLMs internally represent numbers, uncovering linear subspace structures \cite{el2025geometry,zhu2025language}, sublinear spacing patterns \cite{kadlvcik2025pre}, and intrinsic limitations of continuous numerical embeddings \cite{davies2025language}.
NumeroLogic~\cite{schwartz2024numerologic} introduced a simple yet powerful encoding, which prefixes numbers with their digit-length (e.g., \{2:42\}). This format embeds structural information and implicitly prompts the model to recognize place values in arithmetic tasks. However, it mainly modifies the textual encoding format without changing the model architecture, and still relies on discrete token prediction for numerical generation.
SafeAuto \cite{zhang2025safeauto} attempts to address this by introducing a Position-Dependent Cross-Entropy (PDCE) loss, which softens token-level supervision to better respect place-value semantics, but challenges remain—particularly in numerical reasoning tasks such as arithmetic carry-over (e.g., ``9.9'' vs ``10.0'').
xVal~\cite{golkar2023xval} embodies numbers by replacing traditional discrete tokens with a unified ``[NUM]'' token, whose embedding is scaled by the actual number. However, this magnitude-scaling design is less straightforward to directly plug into pretrained pre-norm LLMs/MLLMs, where LayerNorm/RMSNorm may attenuate numerical information encoded only by embedding amplitude. Moreover, in its standard formulation, xVal uniformly replaces numerical values with [NUM], whereas some digit-bearing expressions in driving prompts may function as lexical or template-level text rather than quantities for numerical reasoning.
Our proposed method goes further by treating driving-relevant numerical quantities as a dedicated modality through a learnable number projector and a number head, improving both numerical precision and inference efficiency.




\section{Methodology}
\subsection{Overview}
\begin{figure*}[t]
  \centering
  \includegraphics[width=\linewidth]{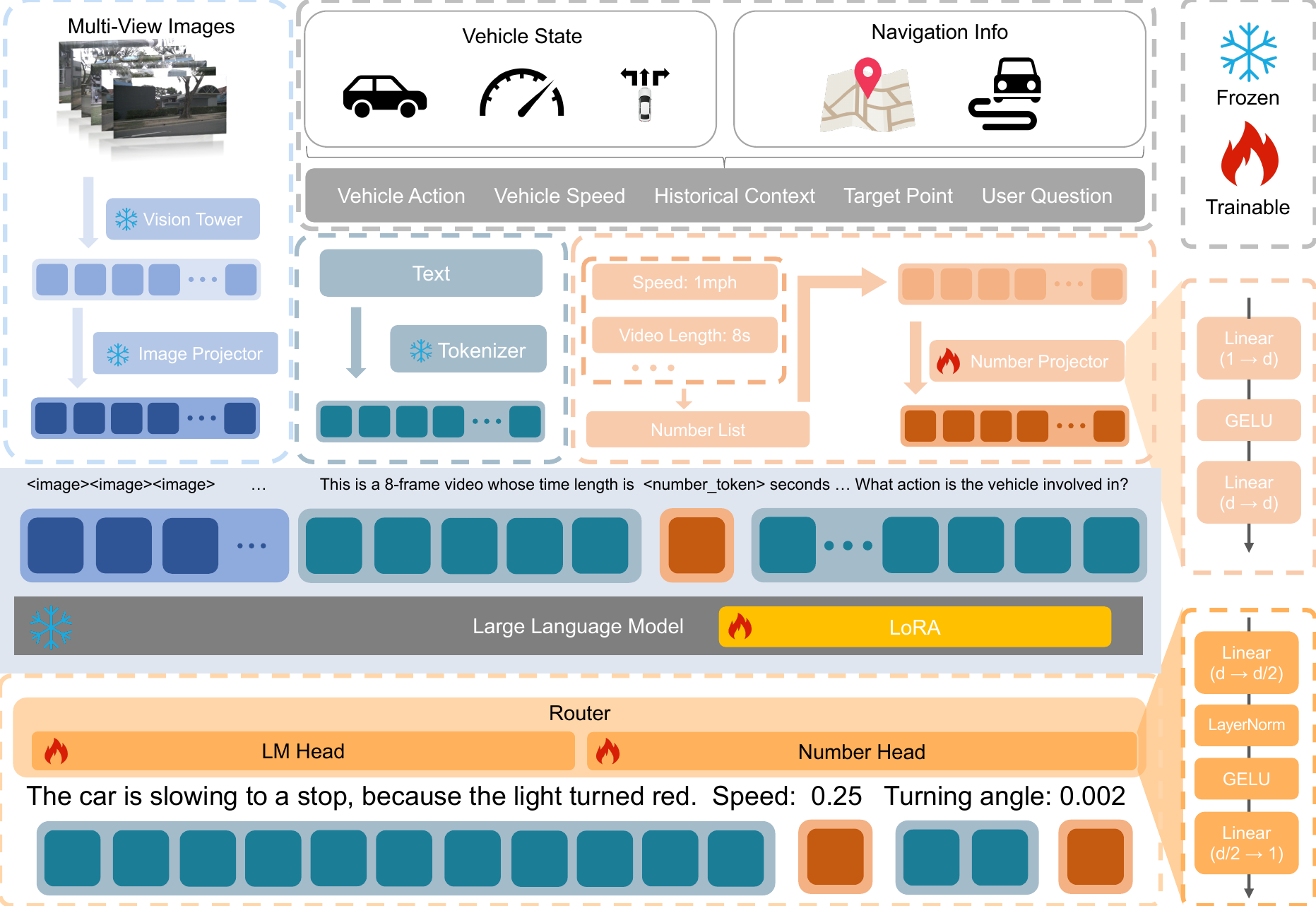}
  
  \caption{DriveCode overview. Our proposed approach consists of three parts: image projection, text tokenization and number projection. The images are first encoded by a vision tower and projected into the language embedding space via an image projector. In parallel, textual descriptions and instructions are tokenized. The third part is the main contribution of our work: continuous numerical signals are vectorized through a dedicated number projector to form aligned numerical tokens. These visual, textual, and numerical tokens are concatenated into a unified sequence and processed by an LLM for further training and inference.}
  \label{diagram}
\end{figure*}

DriveCode is an LLM-based end-to-end autonomous driving framework. It focuses on numerical representation learning via a dedicated number projector and a number head, which are designed to explicitly model and reason over continuous numerical signals related to driving. The overall architecture is shown in Fig. \ref{diagram}.

\subsection{Data Preprocessing}
\label{data_preprocessing}
To enable continuous numerical encoding designed in our work, we implement a preprocessing pipeline across all data samples. This process identifies and extracts numbers from raw text using regular expressions. Only numbers that correspond to truly meaningful physical quantities or control-related signals are converted, whereas descriptive or system-level constants (e.g., the number of camera views) are retained in their original textual form. Each identified number is replaced by a unified special token \textless number\_token\textgreater, which serves as a placeholder for numerical embedding injection. Simultaneously, the original numbers presented in textual form are extracted and converted into floating-point format, then stored in an ordered list that strictly aligns with the sequence of the placeholder \textless number\_token\textgreater. Such an ordered list is maintained for each multi-turn dialogue. 
For example, a descriptive sentence ``The video length is 8 seconds. There are 5 vehicles ahead within a distance of 10.5 meters'' is transformed into ``The video length is \textless number\_token\textgreater \ seconds. There are 5 vehicles ahead within a distance of \textless number\_token\textgreater \ meters'', accompanied with the number list [8.0, 10.5].
Here, the object count ``5'' is kept as text because it is a descriptive count, while video length and distance are converted because they are continuous physical quantities.

\subsection{Model Structure}

DriveCode builds upon LLaVA-NeXT \cite{liu2024llavanext} by introducing a novel number projector. The model architecture comprises: (i) a vision encoder to extract visual features from images, (ii) a vision-language projector for cross-modal alignment between visual tokens and language representations, (iii) a number projector that encodes numbers into the language model’s token embedding space, and (iv) a causal language model for auto-regressive generation.

\textbf{Vision encoder.} Following LLaVA-NeXT\cite{liu2024llavanext}, we adopt SigLIP as the vision backbone. Input images $I$ are processed using AnyRes dynamic resolution strategy: each image is resized to a resolution $(H_g, W_g)$ selected from predefined pinpoints $\{(384, 768), (768, 384), (768, 768), \dots\}$ to preserve aspect ratios, and then split into $336 \times 336$ local patches plus a global view. Visual features are extracted as $F_{global} = \Phi(I_{global})$ and $F_{local}^{(i)} = \Phi(P_i)$, where $\Phi$ denotes the SigLIP encoder and $F \in \mathbb{R}^{h \times w \times d_v}$ with $d_v=1152$. A learnable ``newline" token $t_{nl}$ is inserted between patch rows to maintain spatial structure. The final visual representation $H_{vis}$ concatenates projected global and local features via a two-layer MLP multimodal projector $\mathcal{P}$:

\begin{equation}
\begin{aligned}
H_{vis} =\;& \mathcal{P}(F_{global}) \\
&\oplus \left[ \mathcal{P}(F_{local}^{(1)}) \oplus t_{nl} \oplus \dots \oplus \mathcal{P}(F_{local}^{(N)}) \right].
\end{aligned}
\end{equation}

The language model jointly processes visual embeddings, textual embeddings, and numerical embeddings. The following subsection provides a detailed description of the number projector and its integration with image and text.

\subsection{Proposed Encoding Scheme}

\textbf{Number representation.}
Following the preprocessing described previously in Section \ref{data_preprocessing}, each multi-turn dialogue in datasets contains $N$ placeholder tokens \textless number\_token\textgreater \ with a corresponding ordered list of numbers
\begin{equation}
\mathbf{x} = [x_1, x_2, \ldots, x_N], \quad x_k \in \mathbb{R},
\end{equation}
where the order of $x_k$ strictly matches the order of \textless number\_token\textgreater \ in text sequence. 
We denote the set of positions of numeric placeholders in the token sequence as $\mathcal{N}$.

\textbf{Placeholder index alignment.}
After text tokenization, each \textless number\_token\textgreater \ is assigned a placeholder index denoted as NUMBER\_TOKEN\_INDEX. These indices take negative values to explicitly mark the positions of numerical tokens within the input sequence. Similarly, image placeholders \textless image\textgreater \ are assigned to IMAGE\_TOKEN\_INDEX, yielding the set of visual placeholder positions. We denote the set of positions of visual placeholders as $\mathcal{V}$.

\textbf{Number projector.}
Each number $x_k$ is encoded into the language hidden space via number projector:
\begin{equation}
\mathbf{e}^{(k)}_{\text{num}} = g_{\phi}(x_k) \in \mathbb{R}^{d}
\end{equation}
where $g_{\phi}: \mathbb{R} \to \mathbb{R}^d$ is a two-layer MLP with GELU activation. The first layer projects the number to the hidden dimension $d$, followed by GELU non-linearity, and the second layer refines the representation. This architecture ensures that 1D numbers are expanded into high-dimensional embeddings compatible with the language model.

\textbf{Modality feature integration.}
Let $\mathbf{E}(\cdot)$ denote the standard text encoder, and $\mathbf{f}_{\text{vis}}$ represent the sequence of visual patch features projected to $\mathbb{R}^{d}$ by the image projector. The final input embedding is constructed by inserting modality features into their original placeholder positions:

\begin{equation}
\mathbf{h}^{(0)}_i =
\begin{cases}
\mathbf{f}^{(j)}_{\text{vis}}, & i \in \mathcal{V} \ \text{(the $j$-th visual placeholder)},\\
\mathbf{E}(u_i), & i \notin \mathcal{V} \cup \mathcal{N},\\
\mathbf{e}^{(k)}_{\text{num}}, & i \in \mathcal{N} \ \text{(the $k$-th numeric placeholder)}
\end{cases}
\end{equation}
Here, $u_i$ denotes the token id at position $i$ when this position corresponds to an ordinary text token.

The resulting sequence 
$\mathbf{H}^{(0)}=[\mathbf{h}^{(0)}_1,\ldots,\mathbf{h}^{(0)}_{L}]$
is then fed into the language model. 
Here, $L$ denotes the length of the final multimodal input sequence after inserting visual and numerical embeddings.
This sequence allows the model to jointly attend over visual, textual, and numerical features
while preserving alignment between each \textless number\_token\textgreater \ and its associated number.

\textbf{Number head.}
The number head consists of a linear projection that reduces the hidden dimension to $d/2$, followed by LayerNorm and GELU activation, and a final linear layer that outputs a scalar for numerical regression.\par

\begin{figure}[t]
    \centering
    \includegraphics[width=1\linewidth]{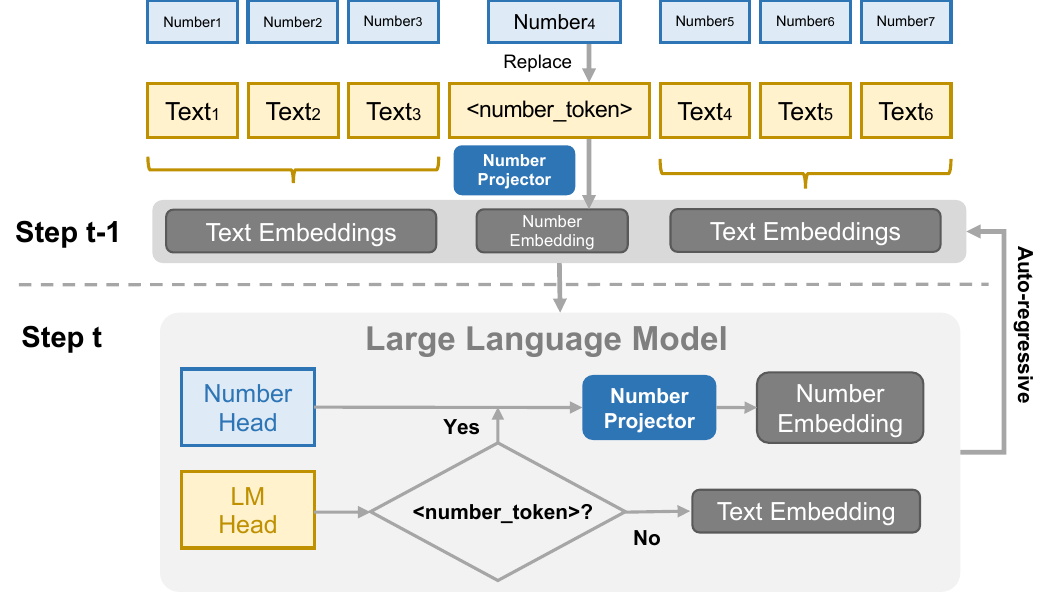}
    \caption{Parallel autoregressive generation of text and numbers. Numbers are projected and fed into the next step without conversion to text embeddings.}
    \label{fig:brief_two_head_intruction}
    
\end{figure}

\textbf{Collaboration of number head and LM head.}
At each step, the LLM produces hidden states that are fed in parallel to a LM head and a number head. When the LM head generates a \textless number\_token\textgreater, the number head's predicted number is passed through the number projector to produce an embedding, which is directly used as the input for the next auto-regressive step. A visual illustration is in Fig. \ref{fig:brief_two_head_intruction}. 
This design better aligns numerical generation with the autoregressive paradigm of LLMs. Unlike fixed-format waypoint heads such as SimLingo~\cite{renz2025simlingo}, DriveCode feeds each generated scalar back through the number projector, allowing later predictions to condition on previous numerical outputs; this is why DriveCode requires both numerical output regression and input-side numerical projection.

\subsection{Loss Function}


DriveCode is trained 
by combining next-token prediction over the textual vocabulary and continuous regression of numerical quantities, including control signals and trajectory waypoints. 
Specifically, when the language model predicts a \textless number\_token\textgreater\ at position $t$, the number head regresses the corresponding number from the hidden state at position $t{-}1$, i.e., the same hidden state used by the LM head for next-token prediction, creating an alignment between discrete token prediction and number regression.

\paragraph{Textual Loss} 
The textual loss is defined as the standard cross-entropy loss:
\begin{equation}
\mathcal{L}_{\text{text}} = - \sum_{i=1}^{L-1} \log p\bigl(y_{i+1} \mid y_{\le i},\, \text{vision},\, \text{numbers}\bigr)
\end{equation}

where positions marked by IGNORE\_INDEX are excluded from the loss summation, i.e., they do not contribute to the computed loss or gradients.

\paragraph{Numerical Loss}
During training, the model predicts continuous numbers at positions immediately preceding each \textless number\_token\textgreater \ in the sequence.
Let $\mathcal{I}=\{i_1,\ldots,i_M\}$ denote the positions of \textless number\_token\textgreater.
For each $i_m \in \mathcal{I}$, the prediction is obtained as:
\begin{equation}
\hat{x}_m = r_{\psi}(\mathbf{h}_{i_m-1})
\end{equation}
where $\mathbf{h}_{i_m-1}$ is the decoder hidden state at position $i_m-1$, and $r_{\psi}$ is the number head described above.
The predicted numbers correspond to either control signals or trajectory waypoints, depending on the task specification.

\textbf{Control signals.}
For scalar quantities such as speed or acceleration, we apply an $\ell_1$ regression loss:
\begin{equation}
\mathcal{L}_{\text{scalar}} = \frac{1}{M_s}\sum_{m=1}^{M_s} |\hat{x}_m - x_m|
\end{equation}
We use $\ell_1$ loss for scalar outputs because it is less sensitive to occasional large residuals, which helps stabilize joint training with the textual next-token prediction loss.

\textbf{Trajectory waypoints.}
For trajectory prediction, we treat the consecutive positions $x_{2t-1}$ and $x_{2t}$ as the waypoints $\hat{\mathbf{p}}_t$, which are supervised using an $\ell_2$ distance:

\begin{equation}
\mathcal{L}_{\text{traj}} = \frac{1}{T}\sum_{t=1}^T \lVert \hat{\mathbf{p}}_t - \mathbf{p}_t \rVert_2
\end{equation}
For waypoint vectors, we use the non-squared $\ell_2$ distance so that the optimization mainly follows the geometric direction of the waypoint error.

\section{Experiments}
\subsection{Datasets}
We evaluate DriveCode on three autonomous driving datasets with different characteristics:

\textbf{DriveGPT4} dataset is derived from the BDD-X dataset and augmented with LLM-generated visual instructions. It includes approximately 20K video clips with control signals, action descriptions, and justifications. Each clip is paired with corresponding Q\&A labels.

\textbf{DriveGPT4-V2} dataset is collected from CARLA simulator using rule-based driving policies, providing a controlled environment for evaluating numerical prediction accuracy in synthetic scenarios, including decision outputs such as steering angle, waypoints, and route point coordinates.

\textbf{OmniDrive} dataset is derived from the nuScenes dataset with OpenLane-v2 annotations. It generates large-scale Q\&A pairs through simulated trajectories and counterfactual reasoning, covering diverse driving behaviors and rule checks. We specifically extracted the trajectory prediction subset to focus on numerical understanding and reasoning capabilities. \par

\begin{figure}[t]
  \centering
  \includegraphics[width=\linewidth]{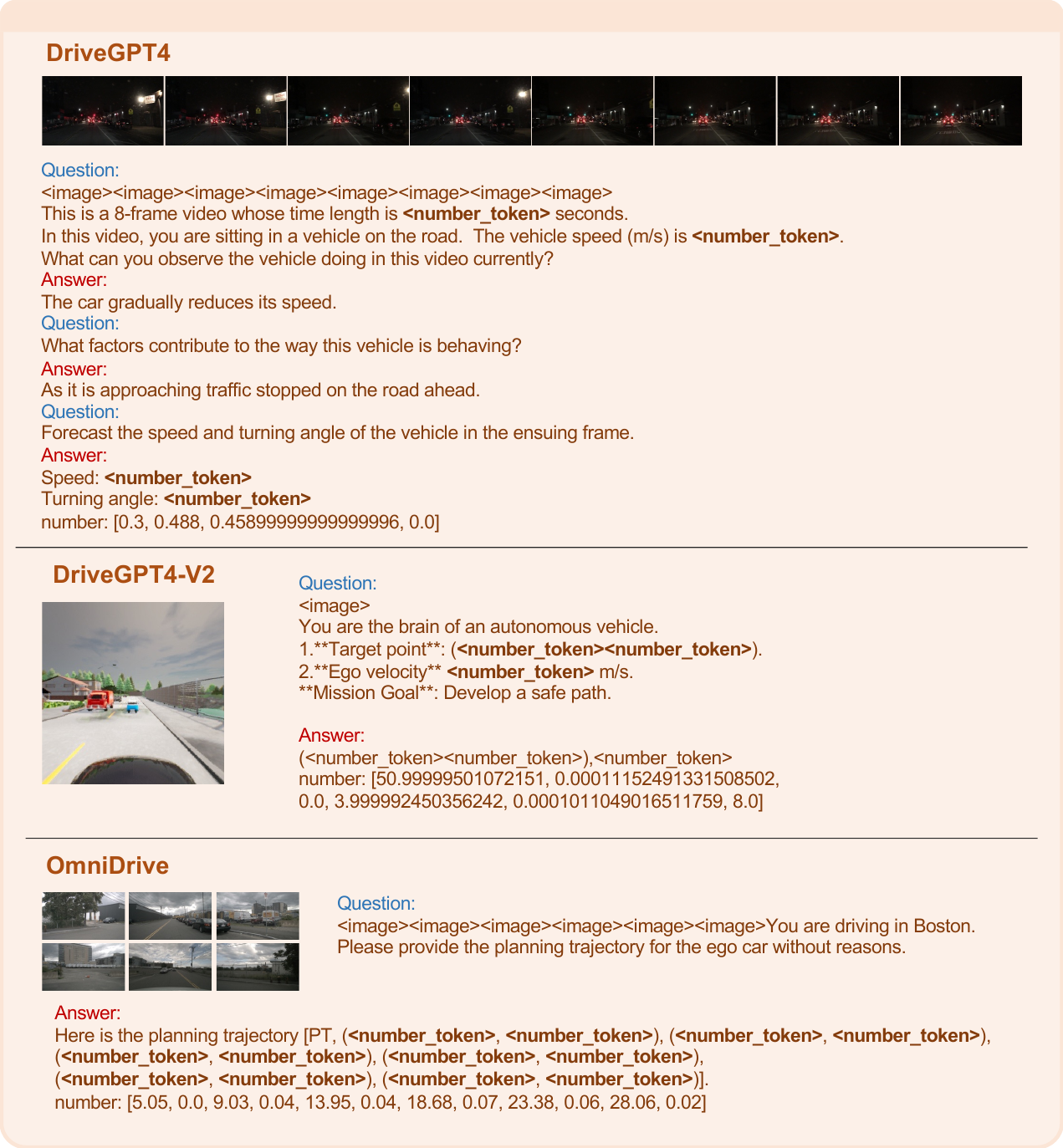}
  \caption{Examples of three datasets. All numbers in these datasets are replaced with \textless number\_token\textgreater.}
  \label{dataset}
\end{figure}

Fig. \ref{dataset} presents examples from the three datasets, demonstrating their input formats and corresponding Q\&A pairs.
\vspace{-8pt}
\subsection{Training Settings}
For DriveGPT4 and DriveGPT4-V2 datasets, DriveCode was trained using PyTorch distributed data parallel on a single node with 8 NVIDIA A100 GPUs. Experiments were conducted for 10 epochs with mixed-precision training (BF16) enabled. We adopted the LLaVA-NeXT training pipeline with a SigLIP-based vision tower and an MLP-based multimodal projector, and optimized the model using the AdamW optimizer with a cosine learning rate schedule and a warmup ratio of 0.03. Gradient checkpointing and DeepSpeed ZeRO-2 were employed to reduce memory consumption. The maximum sequence length was set to 32768 tokens, and training was performed with a per-device batch size of 4 and gradient accumulation steps of 1. The DriveGPT4 dataset contains 16,381 training samples and 2,119 test samples, while DriveGPT4-V2 consists of 152,876 training samples and 26,929 test samples.

For the OmniDrive dataset, DriveCode was trained using PyTorch distributed data parallel on a single node with 2 NVIDIA A100 GPUs. Training was performed with a per-device batch size of 2 and gradient accumulation steps of 1. Other training settings follow those used for DriveGPT4 and DriveGPT4-V2. The OmniDrive dataset contains 21,516 training samples and 5,119 test samples.

\subsection{Evaluation Metrics}

We evaluate DriveCode on both trajectory prediction accuracy and control prediction quality.

\paragraph{Trajectory L2 Distance}
For trajectory prediction, we measure the Euclidean distance between the predicted and ground-truth waypoints. For a 2D target point $\mathbf{p}=[p_x,p_y]$ and prediction $\hat{\mathbf{p}}=[\hat{p}_x,\hat{p}_y]$, the point error is:
\begin{equation}
\mathcal{E}_{\text{point}} = \lVert \hat{\mathbf{p}}-\mathbf{p} \rVert_2
\end{equation}
For multi-step trajectories, we report the mean L2 error across waypoints.

\paragraph{Steering Direction Error}
We compute an angle-based error derived from the predicted 2D direction. We convert the 2D vector into a heading angle in degrees:
\begin{align}
\hat{\theta} &= -\operatorname{deg}\bigl(\arctan2(-\hat{p}_y,\hat{p}_x)\bigr), \\
\theta &= -\operatorname{deg}\bigl(\arctan2(-p_y,p_x)\bigr)
\end{align}

We report the average absolute heading error:
\begin{equation}
\mathcal{E}_{\theta} = \frac{1}{N}\sum_{n=1}^{N} \lvert \hat{\theta}_n-\theta_n \rvert
\end{equation}

\paragraph{Speed Error}
Let $v$ and $\hat{v}$ be the ground-truth and predicted speed. We report the average absolute speed error:
\begin{equation}
\mathcal{E}_{\text{speed}} = \frac{1}{N}\sum_{n=1}^{N} \lvert \hat{v}_n-v_n \rvert
\end{equation}

In our implementation, heading, point, and speed errors are computed \emph{separately} and reported individually, matching the evaluation script. Concretely, for a per-sample error sequence $\{e_n\}$ (e.g., $e_n=\hat{\theta}_n-\theta_n$ for heading, $e_n=\lVert \hat{\mathbf{p}}_n-\mathbf{p}_n\rVert_2$ for points, or $e_n=\hat{v}_n-v_n$ for speed), we compute the normalized L2 norm:
\begin{equation}
\bar{e} = \frac{\lVert [e_1,\ldots,e_N] \rVert_2}{N}
\end{equation}

\subsection{Comparison Experiments}

We compare DriveCode against several baselines that share the same backbone and training recipe, but differ in how numerical information is handled. 
(i) \textbf{ADAPT}: a task-specific action-aware driving captioning model \cite{jin2023adapt}, which generates action descriptions and justifications but does not explicitly model or inject numbers; 
(ii) \textbf{DriveGPT4}: numbers remain as standard text tokens produced by tokenizer; 
(iii) \textbf{xVal}: numbers are represented using scaled embeddings following xVal \cite{golkar2023xval}; 
(iv) \textbf{DriveCode}: each numeric expression is replaced by \textless number\_token\textgreater, and the aligned numbers are injected via the number projector at the corresponding positions.
The detailed results are in Table \ref{tab:control_compare},\ref{tab:traj_l2_omnidrive} and 
\ref{tab:control_compare2}.




\begin{table*}[h]
  \caption{Quantitative results of control signals prediction on the whole DriveGPT4 testing dataset. RMSE denotes root mean square error. $A_{\delta}$ denotes the percentage of samples with absolute error within threshold $\delta$.}
  \renewcommand\tabcolsep{10.1pt} 
  \small
  \label{tab:control_compare}
  \centering
\begin{tabular}{lcccccccccc}
\toprule
    \multirow{2}{*}{Method} & \multicolumn{5}{c}{Speed (m/s)}            & \multicolumn{5}{c}{Turning angle (degree)}      \\ \cmidrule(l){2-6} \cmidrule(l){7-11}  
 & RMSE$\downarrow$  & $A_{0.1}\uparrow$  & $A_{0.5}\uparrow$ & $A_{1.0}\uparrow$ & $A_{5.0}\uparrow$  & RMSE$\downarrow$  & $A_{0.1}\uparrow$  & $A_{0.5}\uparrow$ & $A_{1.0}\uparrow$ & $A_{5.0}\uparrow$\\ 
    \midrule
    ADAPT & 3.02 & 9.56 & 24.77 & 37.07 & 90.39 & 11.98 & 27.93& 66.83 & 75.13 & 89.45\\
    
    DriveGPT4 & 1.30& \textbf{30.09} &60.88 &79.92 &98.44& 8.98 &\textbf{59.23} & \textbf{72.89} &79.59 &\textbf{95.32} \\

    xVal & 1.13& 26.58 &63.46 &82.53 &99.10 & 8.78 &56.99 & 72.89 &80.08 &93.20 \\
    \midrule
    
    DriveCode & \textbf{1.08}& 27.50 &\textbf{64.60} &\textbf{82.99}&\textbf{99.10}& \textbf{7.71} &57.18 & 72.54 &\textbf{80.25} & 93.71\\
    \bottomrule
  \end{tabular}

\end{table*}

\begin{table}[H]
\centering
\setlength{\abovecaptionskip}{4pt}
\setlength{\belowcaptionskip}{0pt}
\renewcommand{\arraystretch}{1.0}
\renewcommand{\tabcolsep}{4.2pt}
\caption{Comparison of Text and DriveCode on trajectory L2 and text answering. (OmniDrive)}
\label{tab:traj_l2_omnidrive}
\begin{tabular}{l c c c}
\toprule
Method & L2 Error (m)$\downarrow$ & {CIDEr}$\uparrow$ & {BLEU4}$\uparrow$ \\
\midrule
Text & 3.0797 & {2.2009} & {26.6945} \\
DriveCode (Ours) & \textbf{2.8274} & {\textbf{2.3829}} & {\textbf{27.3690}} \\
\bottomrule
\end{tabular}
\end{table}

\vspace{-12pt}

\begin{table}[H]
\setlength{\abovecaptionskip}{4pt}
\setlength{\belowcaptionskip}{0pt}
\renewcommand\arraystretch{1.0}
\renewcommand\tabcolsep{4.2pt}
\centering

\begin{threeparttable}
\caption{Comparison of xVal and DriveCode on control signals prediction. (DriveGPT4-V2)}
\label{tab:control_compare2}
\begin{tabularx}{\columnwidth}{l *{3}{>{\centering\arraybackslash}X}}
\toprule
Method &
\makecell[c]{Theta Error \\ (degree)$\downarrow$} &
\makecell[c]{Point Error \\ (L2, m)$\downarrow$} &
\makecell[c]{Speed Error \\ (m/s)$\downarrow$} \\
\midrule

xVal     & 0.07409 & 0.01166 & 0.02162 \\
DriveCode (Ours) & \textbf{0.07377} & \textbf{0.01137} & \textbf{0.02131} \\
\bottomrule
\end{tabularx}
\end{threeparttable}
\end{table}

Overall, DriveCode achieves the best RMSE on both speed and turning angle, and improves most accuracy thresholds, particularly at moderate and large tolerances. While DriveGPT4 retains an advantage at the tightest threshold ($A_{0.1}$), DriveCode's lower RMSE indicates smaller average errors overall, highlighting the advantage of explicitly representing numbers as a dedicated modality.

\begin{figure}[t]
    \centering
    \includegraphics[width=1\linewidth]{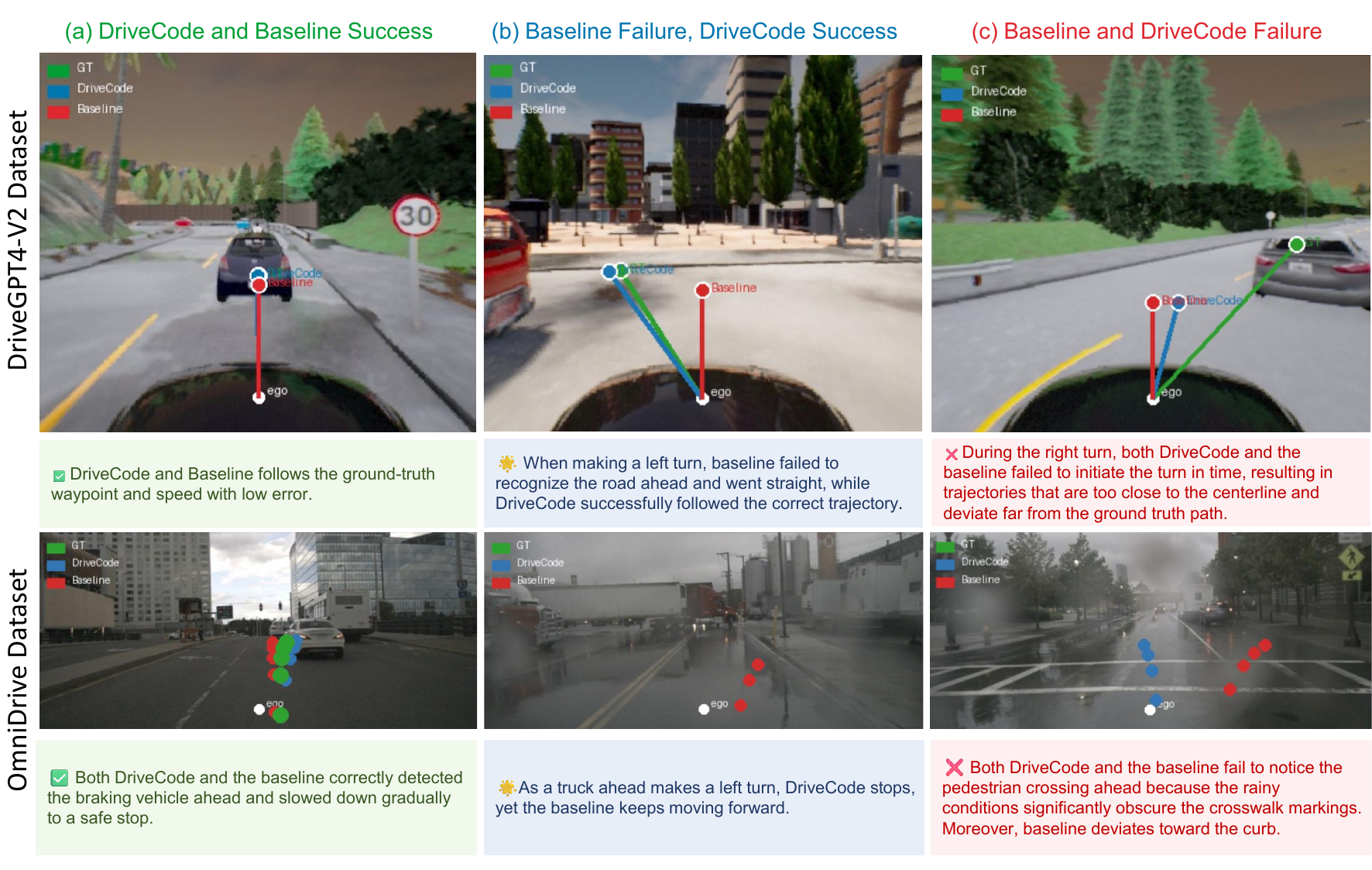}
    \caption{Qualitative comparison of trajectory prediction.}
    \label{fig:traj_vis}
    
\end{figure}

Fig.~\ref{fig:traj_vis} further visualizes representative trajectory prediction cases. Compared with baselines, DriveCode generally produces trajectories 
closer to the ground truth, demonstrating improved numerical modeling for trajectory generation.
\vspace{-6pt}

\subsection{Downstream Driving Evaluation}

To further evaluate whether improved numerical modeling benefits downstream driving performance, we compare DriveCode with SimLingo under the same training setting on the DriveGPT4-V2 dataset.

We further conduct closed-loop driving evaluation on Bench2Drive\cite{jia2024bench} to assess whether improved numerical modeling translates into better driving behavior. Tested on same routes in Bench2Drive, we report driving score for Simlingo and DriveCode. Both methods are evaluated using identical backbone architectures and training settings, differing only in how numerical information is represented.

As shown in Table \ref{tab:closed_loop_compare}, DriveCode achieves consistently better downstream driving performance than SimLingo. In particular, DriveCode obtains higher driving scores, suggesting that explicitly modeling numerical information leads to more accurate control prediction and better driving behavior.

\begin{table}[H]
\centering
\setlength{\abovecaptionskip}{2pt}
\setlength{\belowcaptionskip}{0pt}

\caption{Closed-loop driving evaluation trained on the DriveGPT4-V2 dataset.}
\label{tab:closed_loop_compare}

\begin{tabular}{lcc}
\toprule
Method & Driving Score$\uparrow$  \\
\midrule
SimLingo & 78.02  \\
DriveCode (Ours) & \textbf{86.79}  \\
\bottomrule
\end{tabular}

\end{table}

\vspace{-10pt}
\subsection{Robustness Experiments}

We conduct robustness experiments on normalization strategies, unit variation, numerical magnitude, and sign changes. Detailed experimental settings and results are provided in the supplementary material.

\subsection{Ablation Studies}

We conducted ablation studies to examine the role of numeric information, comparing fully numeric input and output (\textbf{DriveCode}), text input with numeric output (\textbf{Variant}), numeric input with text output (\textbf{variant2}), and text input and output without number projector (\textbf{Text}).

Compared with \textbf{Text}, enabling numeric regression at the output side (\textbf{Variant}) reduces numeric errors, indicating that supervising numbers directly is beneficial. In contrast, \textbf{Variant2}, which only enables numeric conditioning at the input side, achieves limited improvements and even shows performance degradation on several metrics, suggesting that input-side numeric representations alone are insufficient for stable continuous signal prediction. Further enabling numeric conditioning at both the input and output sides (\textbf{DriveCode}) yields the best point and speed accuracy on DriveGPT4-V2 and the best overall performance on DriveGPT4, demonstrating that the aligned numeric stream improves both representation and generation of continuous driving signals. The detailed results are in supplementary material.

\subsection{Efficiency Analysis}

We analyzed the computational efficiency of DriveCode from both training and inference perspectives.
During training, DriveCode introduces a number projector to model numerical information in a continuous embedding space. This module is only applied to numerical tokens and does not significantly increase the overall parameter count or computational complexity. As a result, the training time per epoch of DriveCode is similar to baselines.

During inference, DriveCode demonstrates a slight reduction in inference latency. By encoding numbers into continuous embeddings, the model avoids the computationally expensive process of autoregressively generating multi-token sequences for a single number (e.g., generating ``3", ``.", ``1", ``4" separately). Furthermore, this approach alleviates token-level ambiguity in number reasoning, allowing the model to predict a number in a single decoding step instead of multiple token-level steps. Consequently, this mechanism 
reduces the 
decoding steps required for numerical prediction, resulting in slightly lower inference latency in our experiments.


\subsection{Limitations}
As with most studies, the design of DriveCode is subject to limitations. 
DriveCode depends on reliable extraction and alignment of numbers with  \textless number\_token\textgreater \ occurrences; mismatches, missing numbers, or inconsistent formats can introduce noise. In addition, performance can be sensitive to the scale of numbers and outliers, which motivates careful normalization. Finally, although the numeric channel improves continuous prediction, overall end-to-end driving performance is still bounded by the base LLM's ability.

\section{Conclusion and Future Works}
This paper presents DriveCode, a novel numerical encoding method for LLM-based autonomous driving. By introducing a number projector and a number head, DriveCode maps numbers into the language model's hidden space, alleviating the inherent limitations imposed by discrete tokenization on numerical reasoning. DriveCode jointly processes multimodal inputs, including video frames, textual prompts, and aligned numerical representations, to generate both interpretable textual outputs and high-precision control signals. 
Extensive experiments on the OmniDrive, DriveGPT4, and DriveGPT4-V2 datasets, together with closed-loop simulation evaluations, demonstrate that DriveCode outperforms baselines in trajectory prediction, control signal generation, and downstream driving performance, while ablation studies confirm the effectiveness of explicit numerical encoding and supervision. In the future, we plan to explore multi-scale numerical representations to further enhance the numerical encoding scheme's robustness.
Although this work focuses on autonomous driving as a representative domain with regular numerical formats and physically measurable errors, the proposed numerical interface can be extended to broader text-grounded numerical prediction tasks.

\bibliographystyle{IEEEtran}
\bibliography{mybib}

\clearpage
\twocolumn[
\begin{center}
{\LARGE\bfseries Supplementary Materials for DriveCode: Domain Specific Numerical Encoding for LLM-Based Autonomous Driving\par}
\end{center}
\vspace{1em}
]

\setcounter{table}{0}
\setcounter{figure}{0}
\setcounter{equation}{0}
\renewcommand{\thetable}{A\arabic{table}}
\renewcommand{\thefigure}{A\arabic{figure}}
\renewcommand{\theequation}{A\arabic{equation}}

\label{app:robust_ablation}

\input{robustness_experiments_block}

\input{ablation_studies_appendix_block}

\input{effency}

\begin{figure}[H]
  \centering
  \includegraphics[width=\linewidth]{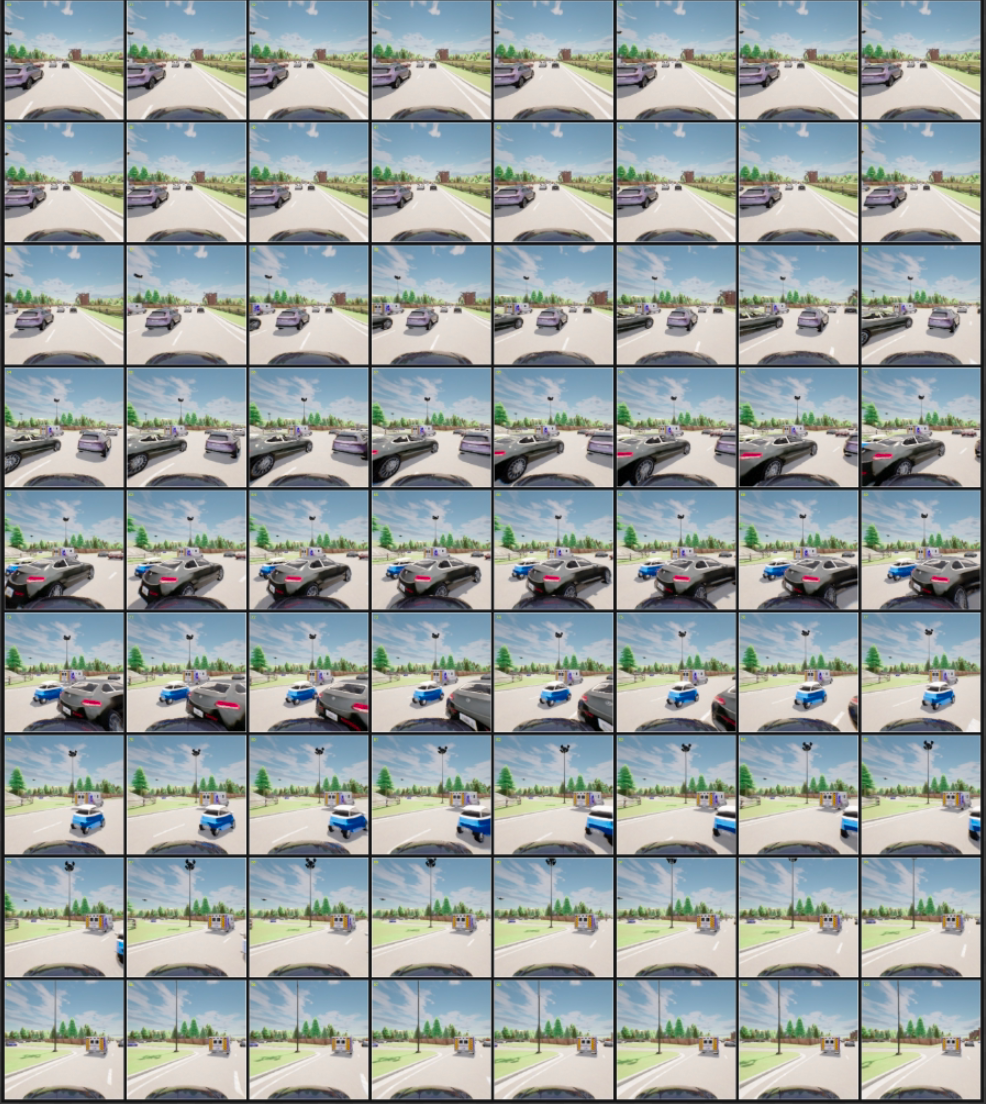}
  \caption{Examples of lane changing with DriveCode: successful lane change, no collision.}
  \label{fig:supplementary_route}
\end{figure}

\end{document}

%% file: robustness_experiments_block.tex

\subsection{Robustness Experiments}

We conduct robustness experiments on normalization strategies, unit variation, numerical magnitude, and sign changes. Decimal variants such as ``3'', ``3.0'', ``3.00'', and ``3e0'' are parsed into the same floating-point value before numerical encoding. Tables \ref{tab:normalization_all}, \ref{tab:robust_unit}, \ref{tab:numerical_robustness}, and \ref{tab:sign_robustness} report the results for normalization, unit variation, magnitude grouping, and sign grouping, respectively.

\begin{table*}[t]
\centering
\begin{threeparttable}
\caption{Comparison of normalization strategies on control signal prediction across DriveGPT4 dataset and DriveGPT4-V2 dataset. \\DriveGPT4 dataset does not include point error evaluation, hence marked as --. Z-score and min-max normalization are applied separately to different numerical categories such as speed, steering angle, and waypoint coordinates.}
\label{tab:normalization_all}
\begin{tabularx}{\textwidth}{l l *{3}{>{\centering\arraybackslash}X}}
\toprule
Dataset & Method &
\makecell[c]{Theta Error (deg)$\downarrow$} &
\makecell[c]{Point Error (L2, m)$\downarrow$} &
\makecell[c]{Speed Error (m/s)$\downarrow$} \\
\midrule

\multirow{3}{*}{\makecell[l]{DriveGPT4\\Dataset}}
& Raw value (DriveCode) & \textbf{7.71} & -- & \textbf{1.08} \\
& Per-type z-score & 8.90 & -- & 1.14 \\
& Per-type min-max & 8.88 & -- & 1.11 \\

\midrule

\multirow{3}{*}{\makecell[l]{DriveGPT4-V2\\Dataset}}
& Raw value (DriveCode) & 0.07377 & 0.01137 & \textbf{0.02131} \\
& Per-type z-score & 0.07276 & \textbf{0.01131} & 0.02164 \\
& Per-type min-max & \textbf{0.07066} & 0.01261 & 0.02229 \\

\bottomrule
\end{tabularx}
\end{threeparttable}
\end{table*}

\begin{table}[H]
\centering

\small
\setlength{\abovecaptionskip}{0pt}
\setlength{\belowcaptionskip}{0pt}
\renewcommand{\arraystretch}{1.0}
\renewcommand{\tabcolsep}{4.2pt}
\begin{threeparttable}
\caption{Unit variation robustness under original and mixed-unit data. speed uses m/s, km/h, mph; angle uses degree and radian (DriveGPT4 dataset)}
\label{tab:robust_unit}
\begin{tabularx}{\columnwidth}{l *{2}{>{\centering\arraybackslash}X}}
\toprule
Setting & Speed RMSE$\downarrow$ & Angle RMSE$\downarrow$ \\
\midrule
Original DriveCode & \textbf{1.08} & \textbf{7.71} \\
Mixed units & 1.34 & 8.89 \\
\bottomrule
\end{tabularx}
\end{threeparttable}
\end{table}

\begin{table*}[t]
\centering

\caption{
Numerical magnitude robustness analysis on the DriveGPT4 dataset.
Samples are grouped according to the magnitude of ground-truth values
using the 33.3\%, 66.7\%, and 95\% quantiles.
}
\label{tab:numerical_robustness}

\footnotesize
\setlength{\tabcolsep}{4pt}

\begin{tabularx}{\textwidth}{
l
>{\centering\arraybackslash}p{2.0cm}
X
X
X
}
\toprule
\textbf{Task} &
\textbf{GT Value Range} &
\textbf{DriveCode} &
\textbf{Variant} &
\textbf{Text-only} \\
\midrule

\multirow{3}{*}{Turning Angle}

& [0, 0.26)
& \textbf{1.1416} (95\% CI: 0.7626, 1.4807; N=1404)
& 1.4035 (95\% CI: 0.9880, 1.7955; N=1403)
& 1.5338 (95\% CI: 1.1637, 1.8924; N=1401) \\

& [0.26, 6.98)
& \textbf{2.1996} (95\% CI: 1.9950, 2.4145; N=608)
& 2.7438 (95\% CI: 2.1821, 3.3726; N=608)
& 2.7683 (95\% CI: 2.4180, 3.1255; N=607) \\

& [6.98, $\infty$)
& \textbf{34.1333} (95\% CI: 23.0185, 44.7093; N=104)
& 37.6696 (95\% CI: 25.9657, 48.5228; N=106)
& 38.7174 (95\% CI: 26.9667, 50.4053; N=107) \\

\midrule

\multirow{4}{*}{Speed}

& [0, 1.44)
& \textbf{0.6544} (95\% CI: 0.4411, 0.8653; N=705)
& 0.7037 (95\% CI: 0.4835, 0.9238; N=705)
& 0.8359 (95\% CI: 0.5740, 1.0823; N=705) \\

& [1.44, 7.54)
& \textbf{1.1763} (95\% CI: 1.0438, 1.3050; N=705)
& 1.2319 (95\% CI: 1.0848, 1.3835; N=706)
& 1.3251 (95\% CI: 1.1534, 1.4963; N=706) \\

& [7.54, 15.70)
& 1.3535 (95\% CI: 1.1455, 1.5731; N=600)
& \textbf{1.3422} (95\% CI: 1.1592, 1.5386; N=600)
& 1.4156 (95\% CI: 1.2430, 1.6002; N=598) \\

& [15.70, $\infty$)
& \textbf{1.0292} (95\% CI: 0.7689, 1.3121; N=106)
& 1.1536 (95\% CI: 0.8704, 1.4487; N=106)
& 1.3385 (95\% CI: 0.9583, 1.6979; N=106) \\

\bottomrule
\end{tabularx}
\end{table*}

\begin{table*}[t]
\centering

\caption{
Sign robustness analysis on the DriveGPT4 dataset.
Samples are grouped into negative, near-zero, and positive intervals
according to the sign of ground-truth values.
Lower is better. Best results are in \textbf{bold}.
}
\label{tab:sign_robustness}

\footnotesize
\setlength{\tabcolsep}{4pt}

\begin{tabularx}{\textwidth}{
l
>{\centering\arraybackslash}p{2.2cm}
X
X
X
}
\toprule
\textbf{Task} &
\textbf{GT Value Range} &
\textbf{DriveCode} &
\textbf{Variant} &
\textbf{Text-only} \\
\midrule

\multirow{3}{*}{theta}
& $(-\infty, -0.05)$
& 9.3818 (95\% CI: 6.4341, 12.2534; N=421)
& \textbf{9.2308} (95\% CI: 6.2741, 12.1288; N=421)
& 9.8490 (95\% CI: 6.6775, 12.6932; N=422) \\

& $[-0.05, 0.05]$
& \textbf{1.1889} (95\% CI: 0.7754, 1.5594; N=1251)
& 1.4857 (95\% CI: 1.0493, 1.9053; N=1250)
& 1.6188 (95\% CI: 1.2121, 1.9940; N=1249) \\

& $(0.05, \infty)$
& \textbf{14.0074} (95\% CI: 7.0870, 19.6879; N=444)
& 16.3430 (95\% CI: 9.2816, 22.5797; N=446)
& 16.7208 (95\% CI: 9.4964, 22.9874; N=444) \\

\midrule

\multirow{3}{*}{speed}
& $(-\infty, -0.05)$
& \textbf{3.5897} (95\% CI: 1.6277, 5.1443; N=13)
& 3.6364 (95\% CI: 1.5787, 5.2521; N=13)
& 3.6722 (95\% CI: 1.4158, 5.3855; N=13) \\

& $[-0.05, 0.05]$
& \textbf{0.1735} (95\% CI: 0.1107, 0.2385; N=366)
& 0.2152 (95\% CI: 0.1223, 0.3109; N=366)
& 0.3085 (95\% CI: 0.1404, 0.4671; N=366) \\

& $(0.05, \infty)$
& \textbf{1.1533} (95\% CI: 1.0518, 1.2607; N=1737)
& 1.1882 (95\% CI: 1.0862, 1.2877; N=1738)
& 1.2948 (95\% CI: 1.1820, 1.4103; N=1736) \\

\bottomrule
\end{tabularx}
\end{table*}

%% file: ablation_studies_appendix_block.tex
\subsection{Ablation Studies}

\begin{table}[H]
\setlength{\abovecaptionskip}{0pt}
\setlength{\belowcaptionskip}{0pt}
\renewcommand\arraystretch{1.0}
\renewcommand\tabcolsep{4.2pt}
\centering
\begin{threeparttable}
\caption{Comparison of Variant and DriveCode on control signals prediction (DriveGPT4).}
\label{tab222}
\begin{tabularx}{\columnwidth}{l *{2}{>{\centering\arraybackslash}X}}
\toprule
Method & Theta Error (degree)$\downarrow$ & Speed Error (m/s)$\downarrow$ \\
\midrule
Variant & 8.63 & 1.11 \\
DriveCode (Ours) & \textbf{7.71} & \textbf{1.08} \\
\bottomrule
\end{tabularx}
\end{threeparttable}
\end{table}

\begin{table}[H]
\setlength{\abovecaptionskip}{2pt}
\setlength{\belowcaptionskip}{2pt}
\renewcommand\arraystretch{0.95}
\renewcommand\tabcolsep{3pt}
\centering
\small
\begin{threeparttable}
\caption{Comparison of different numerical encoding variants on control signal prediction (DriveGPT4-V2).}
\label{tab333}
\begin{tabularx}{\columnwidth}{>{\raggedright\arraybackslash}p{2.5cm} *{3}{>{\centering\arraybackslash}X}}
\toprule
Method &
\makecell[c]{Theta Error \\ (degree)$\downarrow$} &
\makecell[c]{Point Error \\ (L2, m)$\downarrow$} &
\makecell[c]{Speed Error \\ (m/s)$\downarrow$} \\
\midrule
Text & 0.07950 & 0.01363 & 0.02231 \\
Variant & \textbf{0.07078} & 0.01140 & 0.02139 \\
Variant2 &
0.08877 &
0.01391 &
0.02264 \\
DriveCode (Ours) & 0.07377 & \textbf{0.01137} & \textbf{0.02131} \\
\bottomrule
\end{tabularx}
\end{threeparttable}
\end{table}

%% file: effency.tex
\subsection{Efficiency Analysis}

\begin{table}[H] 
    \setlength{\abovecaptionskip}{0pt} 
    \setlength{\belowcaptionskip}{0pt} 
    \renewcommand\arraystretch{1.0} 
    \renewcommand\tabcolsep{4.2pt} 
    \centering 
    \begin{threeparttable} 
        \caption{Efficiency comparison between the baseline model, model using text, model using only \textless number\_token\textgreater \ in output, and DriveCode. Tested on DriveGPT4 dataset.} 
        \label{tab:efficiency} 
        \begin{tabularx}{\columnwidth}{l *{2}{>{\centering\arraybackslash}X}} 
            \toprule  
            Method & {Latency(s)$\downarrow$} & {Avg\_time\_per\_sample(s)$\downarrow$} \\  
            \midrule 
            xVal            & 6776.4096 & 3.1979 \\ 
            Text            & 7152.3616 & 3.3769 \\ 
            Variant         & 6763.7911 & 3.1920 \\ 
            \midrule 
            DriveCode (Ours)& \textbf{6737.9131} & \textbf{3.1798} \\  
            \bottomrule 
        \end{tabularx} 
    \end{threeparttable} 
\end{table}